%% file: Main.tex
\documentclass{article}
\pdfoutput=1
\usepackage{microtype}
\usepackage{graphicx}
\usepackage{booktabs} 
\usepackage{comment}
\usepackage{multirow}
\usepackage[flushleft]{threeparttable}
\usepackage[utf8]{inputenc} 
\usepackage[T1]{fontenc}    
\usepackage{hyperref}       
\usepackage{url}            
\usepackage{breakurl}

\usepackage{amsfonts}       
\usepackage{nicefrac}       
\usepackage{subcaption}
\usepackage{algorithm}
\usepackage{algorithmic}
\usepackage{amssymb}
\usepackage{amsmath}
\usepackage[title]{appendix}
\usepackage{bm}

\newlength\myindent
\usepackage[preprint]{sysml2019}

\sysmltitlerunning{Compressing RNNs for IoT devices by 15-38x using Kronecker Products}

\begin{document}

\twocolumn[
\sysmltitle{Compressing RNNs for IoT devices by 15-38x using Kronecker Products}



\sysmlsetsymbol{equal}{*}

\begin{sysmlauthorlist}
\sysmlauthor{Urmish Thakker}{arm1}
\sysmlauthor{Jesse Beu}{arm1}
\sysmlauthor{Dibakar Gope}{arm1}
\sysmlauthor{Chu Zhou}{arm1}
\sysmlauthor{Igor Fedorov}{arm1}
\sysmlauthor{Ganesh Dasika}{arm1}
\sysmlauthor{Matthew Mattina}{arm1}

\end{sysmlauthorlist}

\sysmlaffiliation{arm1}{Arm ML Research Lab}

\sysmlcorrespondingauthor{Urmish Thakker}{urmish.thakker@arm.com}

\sysmlkeywords{Machine Learning, SysML}

\vskip 0.3in

\input{MainPaper/Abstract}
]



\printAffiliationsAndNotice{}  

\input{MainPaper/Introduction}
\input{MainPaper/RelatedWork}

\input{MainPaper/KPRNN}

\input{MainPaper/Results}
\input{MainPaper/Conclusion}
\bibliographystyle{sysml2019}

\newpage
\clearpage
\begin{appendices}
\input{Appendix/Proof.tex}
\input{Appendix/Dataset}

\input{Appendix/KroneckerProduct}

\input{Appendix/KPRNNHyperparameters}
\input{Appendix/HKPRNNHyperparameters}
\end{appendices}

\end{document}

%% file: MainPaper/Abstract.tex
\begin{abstract}
Recurrent Neural Networks (RNN) can be difficult to deploy on resource constrained devices due to their size. As a result, there is a need for compression techniques that can significantly compress RNNs without negatively impacting task accuracy. This paper introduces a method to compress RNNs for resource constrained environments using Kronecker product (KP). KPs can compress RNN layers by $16-38\times$ with minimal accuracy loss. By quantizing the resulting models to 8-bits, we further push the compression factor to $50\times$. We show that KP can beat the task accuracy achieved by other state-of-the-art compression techniques across 5 benchmarks spanning 3 different applications, while simultaneously improving inference run-time. We show that the KP compression mechanism does introduce an accuracy loss, which can be mitigated by a proposed hybrid KP (HKP) approach. Our HKP algorithm provides fine-grained control over the compression ratio, enabling us to regain accuracy lost during compression by adding a small number of model parameters.

\end{abstract}

%% file: MainPaper/Introduction.tex
\section{Introduction}
\label{sec:intro}
Recurrent Neural Networks (RNNs) achieve state-of-the-art (SOTA) accuracy for many applications that use time-series data. As a result, RNNs can benefit important Internet-of-Things (IoT) applications like wake-word detection \cite{zhang2017}, human activity recognition \cite{hammerla2016deep,opp,deepConvLSTM}, and predictive maintenance \cite{pred1,pred2}. IoT applications typically run on highly constrained devices. Due to their energy, power, and cost constraints, IoT devices frequently use low-bandwidth memory technologies and smaller caches compared to desktop and server processors. For example, some IoT devices have 2KB of RAM and 32 KB of Flash Memory \cite{iot1,iot2}. The size of typical RNN layers can prohibit their deployment on IoT devices or reduce execution efficiency \cite{urmtha01RNN}. Thus, there is a need for a compression technique that can drastically compress RNN layers without sacrificing the task accuracy. 

First, we study the efficacy of traditional compression techniques like pruning \cite{han2015deep_compression,suyog,changpinyo} and low-rank matrix factorization (LMF) \cite{DBLP:journals/corr/KuchaievG17,lmf-bad,lmf-good1}. We set a compression target of $15\times$ or more and observe that neither pruning nor LMF can achieve the target compression without significant loss in accuracy. We then investigate why traditional techniques fail, focusing on their influence on the rank and condition number of the compressed RNN matrices. We observe that pruning and LMF tend to either decrease matrix rank or lead to ill-condition matrices and matrices with large singular values. 

To remedy the drawbacks of existing compression methods, we propose to use Kronecker Products (KPs) to compress RNN layers. We refer to the resulting models as KPRNNs. To mitigate losses in accuracy introduced by KP compression, we propose Hybrid Kronecker product RNNs (HKPRNNs), which regain lost accuracy through the addition of a small set of extra model parameters. In the end, we are able to show that our approach achieves SOTA compression on IoT-targeted benchmarks without sacrificing wall clock inference time and accuracy.

%% file: MainPaper/RelatedWork.tex
\section{Related work}
\label{sec:relwork}

KPs have been used in the deep learning community in the past \cite{kron1,kron2}. For example, \cite{kron2} use KPs to compress fully connected (FC) layers in AlexNet. They start with a pre-trained model and use a low rank decomposition technique to find the sum of KPs that best approximate the FC layer. We deviate from \cite{kron2} by using KPs to compress RNNs and, instead of learning the decomposition for fixed RNN layers, we learn the KP factors directly. Our approach introduces a level of flexibility which \cite{kron2} lacks, since our approach is able to search over a much larger space of factors. Additionally, \cite{kron2} does not examine the impact of compression on inference run-time. In \cite{kron1}, KPs are used to stabilize RNN training through a unitary constraint. A detailed discussion of how the present work differs from \cite{kron1} can be found in Section \ref{sec:kprnn}. KPs have also been used for orthogonal projections for dimensionality reduction of input matrices \cite{kronorthogonal,kronorthogonal2}, which is orthogonal to the present work.

The research in neural network (NN) compression can be roughly categorized into 4 topics: pruning \cite{han2015deep_compression,suyog,changpinyo,blockprune}, structured matrix based techniques \cite{circular2,structuredmatrix,clstm,circnn}, quantization \cite{Quant1,Quant2,Quant3,Quant4,Quant5,Quant-hubara,Quant-bengio,Quant6} and tensor decomposition \cite{cp,tucker,tjandra2017compressing,DBLP:journals/corr/KuchaievG17,lmf-bad,lmf-good1}. 
Compression using structured matrices translates into inference speed-up, but only for matrices of size $2048\times 2048$ and larger \cite{NIPS2018_8119} on CPUs or when using specialized hardware \cite{circular1,circular2}. As such, we restrict our comparisons to pruning and tensor decomposition. 

The choice of RNN cell type has a large impact on model size. For example, GRUs \cite{gru} and LSTMs \cite{lstm} have $3 - 4\times$ more parameters than vanilla RNN cells. Therefore, a straightforward  way to compress RNNs is to replace LSTM and GRU cells with lightweight vanilla RNN Cells. However, vanilla RNN Cells are hard to train and can lead to vanishing and exploding gradients. Thus, any work that leads to stable training of RNN cells can potentially compress NNs by $3 - 4\times$.
Various techniques have been proposed to stabilize RNN training \cite{unitary,unitary2,unitary3,unitary4,unitary5,kron1,msr} . 
FastRNN cells \cite{msr} is the current SOTA approach, showing the best accuracy, fastest runtime, and smallest model size for IoT applications when compared to models trained using LSTMs ,GRUs, and RNNs trained using unitary matrix based stabilization techniques \cite{unitary,unitary2,unitary3,unitary4,unitary5}.  
In this paper, we further compress FastRNN benchmarks by $16\times$ using the proposed technique.

%% file: MainPaper/KPRNN.tex
\section{Kronecker Product Recurrent Neural Networks}
\label{sec:kprnn}

\subsection{Background}
\label{sec:background}
Let $A \in \mathbb{R}^{m\times n}$, $B \in \mathbb{R}^{m_1 \times n_1}$ and $C \in \mathbb{R}^{m_2\times n_2}$. Then, the KP between $B$ and $C$ is given by
\begin{equation}
    A = B\otimes C
    \label{eq:kp}
\end{equation}
\[
A=
\left[ {\begin{array}{cccc}
b\textsubscript{1,1}\circ C & b\textsubscript{1,2}\circ C & ... & b\textsubscript{1,$n_1$}\circ C\\
b\textsubscript{2,1}\circ C & b\textsubscript{2,2}\circ C & ... & b\textsubscript{2,$n_1$}\circ C\\
. & . & . & . \\
. & . & . & . \\
b\textsubscript{$m_1$,1}\circ C & b\textsubscript{1,2}\circ C & ... & b\textsubscript{$m_1$,$n_1$}\circ C\\
\end{array} } \right]
\]
where, $m = m_1 \times m_2$, $n = n_1 \times n_2$, and $\circ$ is the hadamard product. The variables B and C are referred to as the Kronecker factors of A. The number of such Kronecker factors can be 2 or more. If the number of factors is more than 2, we can use  \eqref{eq:kp} recursively to calculate the resultant larger matrix. For example, in the following equation - 
\begin{equation}
    W = W1\otimes W2 \otimes W3
    \label{eq:kp2}
\end{equation}
W can be evaluated by first evaluating $W2 \otimes W3$ to a partial result, say $R$, and then evaluating $W = W1 \otimes R$. 
The algorithm to calculate the Kronecker product (KP) of two matrices in Tensorflow is given in Algorithm \ref{alg:kptf} in Appendix \ref{sec:appKP}. 

Expressing a large matrix A as a KP of two or more smaller Kronecker factors can lead to significant compression. For example, $A \in \mathbb{R}^{154 \times 164}$ can be decomposed into Kronecker factors $B \in \mathbb{R}^{11\times 41}$ and $C \in \mathbb{R}^{14\times 4}$.
The result is a $50\times$ reduction in the number of parameters required to store $A$.
Of course, compression can lead to accuracy degradation, which motivates the present work.

\subsection{Prior work on using KP to stabilize RNN training flow}

Vanilla RNNs are known to suffer from vanishing and exploding gradients \cite{rnnhard}, which motivated the introduction of LSTMs and GRUs. However, LSTMs and GRUs have $3-4\times$ more parameters than vanilla RNNs. Jose et al. \cite{kron1} used KP to stabilize the training of vanilla RNN. An RNN layer has two sets of weight matrices - input-hidden and hidden-hidden (also known as recurrent). The input-hidden matrix gets multiplied with the input, while the hidden-hidden (or recurrent) matrix gets multiplied by the hidden vector. Jose et al. \cite{kron1} use Kronecker factors of size $2\times 2$ to replace the hidden-hidden matrices of every RNN layer. Thus a traditional RNN cell, represented by:
\begin{gather}
    h_t = f([W_x\; \;W_h]*[x_t; h_{t-1}]) 
    \label{eq:origrnn}
\end{gather}
is replaced by,
\begin{gather}
    h_t = f([W_x\;\;\; W_{0} \otimes W_{1}...\otimes W_{F-1}]*[x_t; h_{t-1}]) \label{eq:origkp} 
\end{gather}
where $W_x$ (input-hidden matrix) $\in \mathbb{R}^{m\times n}$, $W_h$ (hidden-hidden or recurrent matrix) $\in \mathbb{R}^{m\times m}$, $W_{i} \in \mathbb{R}^{2\times 2}$ for $i \in \lbrace 0,\cdots,F-1\rbrace$, $x_{t} \in \mathbb{R}^{n\times 1}$, $h_{t} \in \mathbb{R}^{m\times 1}$, and $F=log_{2}(m)=log_{2}(n)$. Thus a $256\times 256$ sized matrix is expressed as a KP of 8 matrices of size $2\times 2$. For an RNN layer with input and hidden vectors of size 256, this can potentially lead to $\sim 2\times$ compression (as we only compress the $W_h$ matrix). The aim of Jose et al. \cite{kron1} was to stabilize RNN training to avoid vanishing and exploding gradients. They add a unitary constraint \cite{linalgbook} to these $2\times 2$ matrices, stabilizing RNN training. However, in order to regain baseline accuracy, they needed to increase the size of the RNN layers significantly, leading to more parameters being accumulated in the $W_x$ matrix in \eqref{eq:origkp}. Thus, while they achieve their objective of stabilizing vanilla RNN training, they achieve only minor compression ($<2\times$). 

In this paper, we show how to use KP to compress both the input-hidden and hidden-hidden matrices of vanilla RNN, LSTM and GRU cells and achieve significant compression ($>16\times$). We show how to choose the size and the number of Kronecker factor matrices to ensure high compression rates , minor impact on accuracy, and inference speed-up over baseline on an embedded CPU. Additionally, we show that, in some cases, there is a need for fine-grained control of the compression rate, which we implement using  hybrid KP.

\subsection{KPRNN Layer}

\subsubsection{Choosing the number of Kronecker factors}

A matrix expressed as a KP of multiple Kronecker factors can lead to significant compression. However, deciding the number of factors is not obvious. 

We started by exploring the framework of \cite{kron1}. We used $2\times 2$ Kronecker factor matrices for hidden-hidden/recurrent matrices of GRU layers \cite{gru} of the key-word spotting network \cite{zhang2017}. This resulted in an approximately $2\times$ reduction in the number of parameters. However, the accuracy dropped by 4\% relative to the baseline. When we examined the $2\times 2$ matrices, we observed that, during training, the values of some of the matrices hardly changed after initialization. This behavior may be explained by the fact that the gradient flowing back into the Kronecker factors vanishes as it gets multiplied with the chain of $2\times2$ matrices during back-propagation. In general, our observations indicated that as the number of Kronecker factors increased, training became harder, leading to significant accuracy loss when compared to baseline.
 
 \begin{algorithm}[htb]
   \caption{Implementation of matrix vector product, when matrix is expressed as a KP of two matrices}
   \label{alg:kpmv}
   \textbf{Input}: Matrices $B$ of dimension $m_1 \times n_1$, $C$ of dimension $m_2 \times n_2$ and $x$ of dimension $n \times 1$. $m = m_1\times m_2$, $n = n_1\times n_2$ \\
   \textbf{Output}: Matrix $y$ of dimension $m\times 1$ 
   \begin{algorithmic}[1]
   \STATE $X = reshape(x,n_2,n_1)$ \COMMENT{reshapes the x vector to a matrix of dimension $n_2\times n_1$}
   \STATE $Bt = B.transpose()$
   \STATE $Y = C\times X\times Bt$
   \STATE $y = reshape(Y,m,1)$ \COMMENT{reshapes the y vector to a matrix of dimension $m\times 1$}
\end{algorithmic}
\end{algorithm}
 
Additionally, using a chain of $2\times 2$ matrices leads to significant slow-down during inference on a CPU. For inference on IoT devices, it is safe to assume that the batch size will be one. When the batch size is one, the RNN cells compute matrix vector products during inference. To calculate the matrix-vector product, we need to multiply and expand all of the $2\times 2$ to calculate the resultant larger matrix, before executing the matrix vector multiplication. Referring to \eqref{eq:origkp}, we need to multiply $W_0,..,W_F$ to create $W_h$ before executing the operation $W_h*h_{t-1}$. The process of expanding the Kronecker factors to a larger matrix, followed by matrix-vector products, leads to a slower inference than the original uncompressed baseline. Thus, inference for RNNs represented using \eqref{eq:origrnn} is faster than the compressed RNN represented using \eqref{eq:origkp}. The same observation is applicable anytime the number of Kronecker factors is greater than $2$.  The slowdown with respect to baseline increases with the number of factors and can be anywhere between $2 - 4\times$.
 
However, if the number of Kronecker factors is restricted to two, we can avoid expanding the Kronecker factors into the larger matrix and achieve speed-up during inference. Algorithm \ref{alg:kpmv} shows how to calculate the matrix vector product when the matrix is expressed as a KP of two Kronecker factors. The derivation of this algorithm can be found in \cite{kpmv} and has been reproduced in Appendix \ref{app:kpproofmv} for convenience.  

\subsubsection{Choosing the dimensions of Kronecker factors}

\begin{algorithm}[tb]
   \caption{Finding dimension of Kronecker Factors for a matrix of dimension $m\times n$}
   \label{alg:primefactors}
   \textbf{Input}: $list1$ is the sorted list of prime factors of $m$, $list2$ is the sorted list of prime factors of $n$ \\
   \textbf{Output}: $listA$ - Dimension of the first Kronecker factor. $listB$ - Dimension of the second Kronecker factor
   \begin{algorithmic}[1]
   \STATE function reduceList (inputList)
        \STATE \hskip1.0em temp1 = inputList[0]
        \STATE \hskip1.0em      inputList.del(0) //Delete the element at position zero
        \STATE \hskip1.0em      inputList[0] = inputList[0]*temp1
        \STATE \hskip1.0em      inputList.sort('ascending')
        \STATE \hskip1.0em return inputList
   \STATE
   \STATE list2 = reduceList(list2)
   \STATE list1 =  reduceList(list1).sort('descending')
   \STATE listA = [list1[0],list2[0]]
   \STATE listB = [list1[1],list2[1]]
\end{algorithmic}
\end{algorithm}

A matrix can be expressed as a KP of two Kronecker factors of varying sizes. The compression factor is a function of the size of the Kronecker factors. For example, a $256\times 256$ matrix can be expressed as a KP of $2\times 2$ and $128\times 128$ matrices, leading to a $4\times$ reduction in the number of parameters used to store the matrix. However, if we use Kronecker factors of size $32 \times 8$ and $8 \times 32$, we achieve a compression factor of $128$. In this paper, we choose the dimensions of the factors to achieve maximum compression, as shown in Algorithm \ref{alg:primefactors} 
. The algorithm takes in the prime factors of the dimensions of the input matrix and returns the dimensions of the two Kronecker factor matrices by converting the list of prime factors for each input dimension into the smallest two numbers, whose product will return a value equal to that dimension. Empirically, this gives the point of maximum compression. 

\subsubsection{Compressing LSTMs, GRUs and RNNs using the KP}
KPRNN cells are RNN, LSTM and GRU cells with all of the matrices compressed by replacing them with KPs of two smaller matrices. For example, the RNN cell depicted in \eqref{eq:origrnn} is replaced by: 
\begin{gather}
    KPRNN\;\;cell: h_t = f((W_{1} \otimes W_{2})*[x_t; h_{t-1}])
    \label{eq:kprnn}
\end{gather}
where $x_{t} \in \mathbb{R}^{n\times 1}$, $h_{t} \in \mathbb{R}^{m\times 1}$, $W_{1} \in \mathbb{R}^{m_1\times n1}$, $W_{2} \in \mathbb{R}^{m_2\times n2}$, $m_1\times m_2 = m$ and $n_1\times n_2 = (m+n)$. 
LSTM, GRU and FastRNN cells are compressed in a similar fashion.
Instead of starting with a trained network and decomposing its matrices into Kronecker factors, we replace the RNN/LSTM/GRU cells in a NN with its KP equivalent and train the entire model from the beginning.  

\subsection{Hybrid KPRNN}
\label{sec:hkprnn}

\begin{figure}[htb]
  \begin{center}
  \includegraphics[width=\columnwidth]{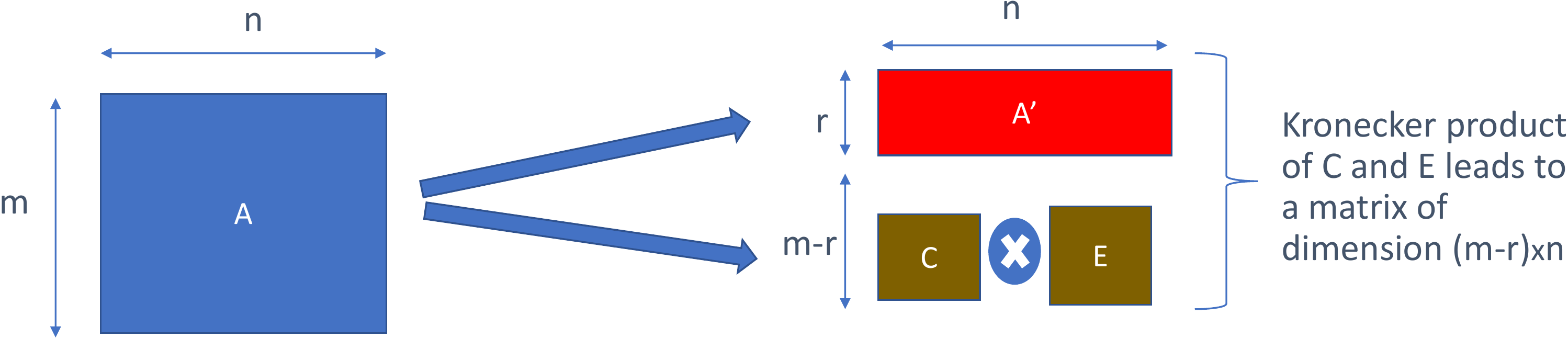}
  \caption{Matrix representation for matrices in a HKPRNN cells}
  \label{fig:hkprnn}
  \end{center}
\end{figure}
\begin{algorithm}[htb]
   \caption{Implementation of Matrix Vector Product, when matrix is expressed as a Hybrid of unconstrained upper part and a lower part created using KP of two matrices}
   \label{alg:hkpmv}
   \textbf{Input}: Matrices $A'$ of dimension $r\times n$, $C$ of dimension $m_1 \times n_1$, $E$ of dimension $m_2 \times n_2$ and $x$ of dimension $n \times 1$. $m_1\times m_2 = (m-r)$, $n = n_1\times n_2$ \\
   \textbf{Output}: Matrix $y$ of dimension $m\times 1$ 
   \begin{algorithmic}
   \STATE $y_{upper} = A'\times x$
   \STATE $X = reshape(x,n_2,n_1)$ \COMMENT{reshapes the x vector to a matrix of dimension $n_2\times n_1$}
   \STATE $At = A.transpose()$
   \STATE $Y_{lower} = B\times X\times At$
   \STATE $y_{lower} = reshape(Y,m,1)$ \COMMENT{reshapes the x vector to a matrix of dimension $(m-r)\times 1$}
   \STATE $y = concat(y_{upper},y_{lower})$
   \end{algorithmic}
\end{algorithm}
\subsubsection{Fine-grained compression rate control}
KP can be an extremely effective compression technique, as will be shown in Section \ref{sec:results-kprnn}. However, sometimes the accuracy loss induced by KP compression may be too large for the technique to be useful. For other compression techniques like pruning \cite{suyog} and LMF \cite{svd}, one way to recover the accuracy is to reduce the amount of compression. This can be done by reducing the sparsity (pruning) of the network or increasing the rank of the matrix (LMF).  These control mechanisms can help regain some of the lost accuracy as they increase the number of parameters in the layer.


\subsubsection{Possible methods to control the amount of compression of KPRNN network}
\label{sec:controlkp}
\paragraph{Method 1:} This paper compresses a network using KP by choosing the point of maximum compression. However, a given matrix can have multiple sets of Kronecker factors which vary in dimensionality. For example, there are multiple ways to express a $256\times 256$ as a KP of two Kronecker factors. Kronecker factors of dimension $2\times 2$ and $128\times 128$ will lead to a compression by a factor of $4\times$, while Kronecker factors of dimension $32 \times 8$ and $8 \times 32$ will lead to a compression by a factor of $128\times$. Apart from these two, there are 6 other compression factors - $2\times, 4\times, 8\times, 16\times, 32\times, 60\times, 102\times$. However, the number of options available to compress a network this way are very limited. For example, if the results after compressing the network by $32\times$ are not satisfactory, the next compression point that can be targeted is $16\times$.

\paragraph{Method 2:} Another way to control the factor by which a KPRNN network is compressed is by increasing the size of the Kronecker factor matrices such that the size of the resultant matrix after KP increases. Increasing the size of the KP leads to an RNN layer with a larger hidden vector. While this might well lead to a valid solution, it removes the possibility of using KPRNN as an in-place replacement in an existing RNN application. This also increases the size of the softmax layers or the subsequent RNN layers that usually follow an RNN layer. For example, referring to \eqref{eq:kprnn}, if we double the size of $W_1$ and $W_2$, the size of the resultant KP will increase by $4\times$. This implies that the size of $h_t$ increases by 4. As a result, the size of the matrices in the subsequent RNN or softmax layer also increase.
 
\subsubsection{HKPRNN}
In order to to achieve fine-grained control of the compression factor achieved by KPRNN networks, while still guaranteeing that the resulting approximation can be an in-place replacement of RNN layers, we propose the Hybrid KP (HKP) mechanism to compress RNNs. We refer to RNN, LSTM and GRU network compressed using this mechanism as, HKPRNN. HKPRNN is inspired from HMD proposed in \cite{thakker2019runtime}. HKP divides a matrix in a NN into two parts, an unconstrained upper part and a lower part created using the KP of two matrices (see Figure \ref{fig:hkprnn}). Thus, the RNN cells in \eqref{eq:origrnn} are replaced by:
\begin{gather}
    HKPRNN Cell: h_t = f([A';(W_{1} \otimes W_{2})]*[x_t; h_{t-1}])
\end{gather}
where $A'\in \mathbb{R}^{r\times (m+n)}$, $W_{1} \in \mathbb{R}^{m_1\times n_1}$, $W_{2} \in \mathbb{R}^{m_2\times n_2}$, $m_1\times m_2 = (m-r)$ and $n_1\times n_2 = (m+n)$. LSTMs and GRUs are compressed in a similar fashion.

By cleverly selecting $r$, we can tune the amount of compression. For example, a $256\times 256$ matrix  can be compressed to $28$ different factors for different values of $r$. 
Thus, controlling the compression using HKPRNN has advantages over both Method 1 and Method 2 described in \ref{sec:controlkp}. It provides more points of compression than Method 1 and does not alter the network architecture as in Method 2. HKP also has accuracy advantages over Method 1 (shown in Section \ref{sec:ablation}). Algorithm \ref{alg:hkpmv} shows how to calculate the matrix vector product without expanding the matrix into its full representation, when a matrix is expressed using the HKP representation.

%% file: MainPaper/Results.tex
\section{Results}
\label{sec:results}

\begin{table*}
\centering
\caption{Benchmarks evaluated in this paper. These benchmarks represent some of the key applications in the IoT domain. We cover a wide variety of applications and RNN cell types.}
\label{tab:benchmarks}
\begin{tabular}{|p{0.28\columnwidth}|p{0.28\columnwidth}|p{0.28\columnwidth}|p{0.28\columnwidth}|p{0.28\columnwidth}|p{0.28\columnwidth}|}
\hline
 & MNIST- & USPS-& KWS-& KWS-& HAR1- \\ 
 & LSTM & FastRNN & LSTM & GRU & BiLSTM\\ \hline
\begin{tabular}[c]{@{}l@{}}Application \\ Domain\end{tabular} & \begin{tabular}[c]{@{}l@{}}Image \\ Classification\end{tabular} & \begin{tabular}[c]{@{}l@{}}Image \\ Classification\end{tabular} & \begin{tabular}[c]{@{}l@{}}Key-word \\ spotting\end{tabular} & \begin{tabular}[c]{@{}l@{}}Key-word\\ spotting\end{tabular} & \begin{tabular}[c]{@{}l@{}}Human \\ Activity \\ Recognition\end{tabular} \\ \hline
Reference Paper &  &\cite{msr}  &\cite{zhang2017}  &\cite{zhang2017}  &\cite{hammerla2016deep}  \\ \hline
Cell Type & LSTM & FastRNN & LSTM & GRU & \begin{tabular}[c]{@{}l@{}}Bidirectional \\ LSTM \end{tabular} \\ \hline

Dataset & \cite{mnist} & \cite{usps} & \cite{warden} &\cite{warden} &\cite{opp} \\ \hline Accuracy & 99.40\% & 93.77\% & 92.50\% & 93.50\% & 91.90\% \\ \hline
\#Parameters & 11,450 & 1,856 & 62,316 & 78,090 & 374,468 \\ \hline
\end{tabular}
\vspace{-1.0em}
\end{table*}

\begin{table*}[t]
\vspace{0cm}
\centering
\begin{threeparttable}
\caption{Model accuracy and runtime for our benchmarks before and after compression. The baseline networks are compared to networks with RNN layers in the baseline compressed using KPs, magnitude pruning, LMF, or by scaling the network size (Small Baseline). Each compressed network has fewer RNN parameters than the baseline (size indicated). For each row, best results are indicated in bold. The KP-based networks are consistently the most accurate alternative while still having speed-up over the baseline.}
\label{tab:kprnnnresults}
\begin{tabular}{|l|l|l|l|l|l|l|}
\hline
\multirow{2}{*}{Benchmark Name} &
\multirow{2}{*}{Parameter measured} &
\multicolumn{1}{c|}{} & \multicolumn{4}{c|}{Compression Technique} \\ \cline{3-7}
\multicolumn{1}{|c|}{} & \multicolumn{1}{c|}{} & \multicolumn{1}{c|}{\textit{Baseline}} & \multicolumn{1}{c|}{Small Baseline} & \multicolumn{1}{c|}{\begin{tabular}[c]{@{}c@{}}Magnitude\\ Pruning\end{tabular}} & LMF & KP \\  \hline
\multirow{4}{*}{MNIST-LSTM} & Model Size (KB)\tnote{1} & \textit{44.73} & 4.51 & 4.19 & 4.9 & \textbf{4.05} \\ \cline{2-7} 
\multicolumn{1}{|c|}{} & Accuracy (\%) & \textit{99.40} & 87.50 & 96.49 & 97.40 & \textbf{98.44} \\ \cline{2-7} 
\multicolumn{1}{|c|}{} & Compression factor \tnote{2} & \textit{1}$\times$ & $10\times$ & $16.7\times$ & $13.08\times$ & \textbf{17.6}$\bm{\times}$ \\ \cline{2-7} 
\multicolumn{1}{|c|}{} & Runtime (ms) & \textit{6.3} & 0.7 & \textbf{0.66} & 1.8 & 4.6 \\ \hline
\multicolumn{7}{|l|}{} \\ \hline
\multirow{4}{*}{HAR1-BiLSTM} & Model Size (KB)\tnote{1} & \textit{1462.3} & 75.9 & 75.55 & 76.39 & \textbf{74.90} \\ \cline{2-7} 
 & Accuracy (\%) & \textit{91.90} & 88.84 & 82.97 & 89.94 & \textbf{91.14} \\ \cline{2-7} 
 & Compression factor \tnote{2} & \textit{1}$\times$ & 19.8$\times$ & 28.6$\times$ & 28.1$\times$ & \textbf{29.7}$\bm{\times}$ \\ \cline{2-7} 
 & Runtime (ms) & \textit{470} & \textbf{29.92} & 98.2 & 64.12 & 157 \\ \hline
 \multicolumn{7}{|l|}{} \\ \hline
\multirow{4}{*}{KWS-LSTM} & Model Size (KB)\tnote{1} & \textit{243.4} & 16.3 & 15.56 & 16.79 & \textbf{15.30} \\ \cline{2-7} 
 & Accuracy (\%) & \textit{92.5} & 89.70 & 84.91 & 89.13 & \textbf{91.2} \\ \cline{2-7} 
 & Compression factor \tnote{2} & \textit{1}$\times$ & 15.8$\times$ & 23.81$\times$ & 21.2$\times$ & \textbf{24.47}$\bm{\times}$ \\ \cline{2-7} 
 & Runtime (ms) & \textit{26.8} & \textbf{2.01} & 5.9 & 4.1 & 17.5 \\ \hline
\multicolumn{7}{|l|}{} \\ \hline
\multirow{4}{*}{KWS-GRU} & Model Size (KB)\tnote{1} & \textit{305.04} & 22.62 & - & 22.63 & \textbf{15.01} \\ \cline{2-7} 
 & Accuracy (\%) & \textit{93.5} & 84.53 & - & 90.88 & \textbf{92.3} \\ \cline{2-7} 
 & Compression factor \tnote{2} & \textit{1}$\times$ & 13.48$\times$ & - & 12.45$\times$ & \textbf{38.45}$\bm{\times}$ \\ \cline{2-7} 
 & Runtime (ms) & \textit{67} & \textbf{6} & - & 7.2 & 17 \\ \hline
\multicolumn{7}{|l|}{} \\ \hline
\multirow{4}{*}{USPS-FastRNN} & Model Size (KB)\tnote{1} & \textit{7.25} & 1.98 & 1.92 & 2.04 & \textbf{1.63} \\ \cline{2-7} 
 & Accuracy (\%) & \textit{93.77} & 91.23 & 88.52 & 89.56 & \textbf{93.20} \\ \cline{2-7} 
 & Compression factor \tnote{2} & \textit{1}$\times$ & 4.4$\times$ & 8.94$\times$ & 8$\times$ & \textbf{16}$\bm{\times}$ \\ \cline{2-7} 
 & Runtime (ms) & \textit{1.17} & 0.4 & \textbf{0.375} & 0.28 & 0.6 \\ \hline
\end{tabular}
\begin{tablenotes}
  \item[1] Model size is calculated assuming 32-bit weights. Further opportunities exist to compress the network via quantization and compressing the fully connected softmax layer.
  \item[2] We measure the amount of compression of the LSTM/GRU/FastRNN layer of the network
\end{tablenotes}
\end{threeparttable}
\end{table*}

\paragraph{Other compression techniques evaluated:} We compare networks compressed using KPRNN and HKPRNN with three techniques: 
\begin{itemize}
    \item \textit{Pruning}: We use the magnitude pruning framework provided by \cite{suyog}. While there are other possible ways to prune, recent work \cite{bestprune} has suggested that magnitude pruning provides state-of-the-art or comparable performance when compared to other pruning techniques \cite{varprune,l0prune}. Pruning creates sparse matrices which are stored in a specialized sparse data structure such as compressed sparse row format (CSR). The overhead of traversing these data structures while performing the matrix-vector multiplication can lead to poor inference run-time compared to a baseline network with dense matrices.
    \item \textit{Low-rank Matrix Factorization (LMF)}: LMF ~\cite{DBLP:journals/corr/KuchaievG17} expresses a matrix $A \in \mathbb{R}^{m\times n}$ as a product of two matrices $U \in \mathbb{R}^{m \times d}$ and $V \in \mathbb{R}^{d \times n}$, where $d$ controls the compression factor. 
   \item \textit{Small Baseline}: Additionally, we train a smaller baseline with the number of parameters equal to that of the compressed baseline. The smaller baseline helps us evaluate if the network was over-parameterized. 
\end{itemize}

\paragraph{Training platform, infrastructure, and inference run-time measurement:} \label{sec:trainingPI} We use Tensorflow 1.12 \cite{tensorflow2015-whitepaper} as the training platform and 4 Nvidia RTX 2080 GPUs to train our benchmarks. To measure the inference run-time, we implement the baseline and the compressed cells in C++ using the Eigen library \cite{eigen} and run them on the Arm Cortex-A73 core of a Hikey 960 development board. 

\paragraph{Dataset and data pre-processing:} We evaluate the impact of compression using the techniques discussed in Section \ref{sec:kprnn} on a wide variety of benchmarks spanning applications like key-word spotting, human activity recognition, and image classification.

\begin{itemize}
    \item \textit{Human Activity Recognition}: We use the \cite{opp} dataset for human activity recognition. We split the benchmark into training, validation, and test data using the procedure described in \cite{hammerla2016deep}. 
    For each input in the dataset, a 81 dimensional vector is fed to the network over 77 time steps. 
    \item \textit{Image Classification}: We use the MNIST \cite{mnist} and USPS \cite{usps} dataset for image classification. The USPS dataset consists of 7291 train and 2007 test images while the MNIST dataset consists of 60k training and 10k test images. We split the publicly available training set into 80\% training set and 20\% validation set and use the selected set of hyperparameters on the test set.
    \item \textit{Key-word Spotting}: We use the \cite{warden} dataset for key-word spotting. The entire dataset consists of 65K different samples of 1-second long audio clips of 30 keywords, collected from thousands of people. We split the benchmark into training, validation and test dataset using the procedure described in \cite{zhang2017}.
\end{itemize}

The details regarding input pre-processing for various benchmarks can be found in Appendix \ref{sec:appDataPreProcess}.
\paragraph{Benchmarks:} Table \ref{tab:benchmarks} shows the benchmarks used in this work. The hyperparameters used for baseline networks are discussed in Appendix \ref{sec:appBaselineAlgorithmsAndImplementation}. MNIST-LSTM takes in input of size $28\times 1$ and feeds it to a LSTM layer of size 40 over $28$ time-steps. USPS-FastRNN takes in input of size $16\times 1$ and feeds it to a FastRNN layer of size $32$ over $16$ time-steps. KWS-GRU network feeds in $10\times 1$ dimension input over $25$ time-steps to a GRU layer of size $154$. KWS-LSTM network feeds in $10\times 1$ dimension input over $25$ time-steps to a LSTM layer of size $118$. HAR1-BiLSTM layer feeds in $79\times 1$ dimension input over $81$ time-steps to a bidirectional LSTM layer \cite{blstm} of size $178$. For all of these networks, the RNN layers are followed by a fully-connected softmax layer. However, in this work, we only focus our attention on compressing the RNN layers.
\paragraph{Evaluation Criteria:}\label{sec:evaluate} We evaluate and compare the compressed networks based on the final accuracy of the network on the held out test set. We also measure the run-time (wall clock time taken to execute a single inference) on the Hikey platform and report the speed-up over the baseline. To measure the wall clock inference run-time, we implement these cells using the eigen library and run them on a single A73 core of the Hikey platform.

Together, the two metrics above help us evaluate whether the proposed training technique can help recover accuracy after significant compression without sacrificing any real-time deadlines IoT-targeted applications may have.

\subsection{KPRNN networks}
\label{sec:results-kprnn}

Table \ref{tab:kprnnnresults} shows the results of applying the KP compression technique across a wide variety of applications and RNN cells. As mentioned in Section \ref{sec:kprnn}, we target the point of maximum compression using two matrix factors.
The RNN layer compression factor for each network is reported in Table \ref{tab:kprnnnresults} and is substantial, ranging from 16$\times$ to 38$\times$. The KP compressed networks are compared to the uncompressed baseline and the networks generated when alternative compression techniques are used to achieve the same compression ratio. This allows for a fair comparison of accuracy and run-time across different techniques. We find that KP compressed networks consistently outperform competing methods, while still providing a speed-up over the baseline network. 

The USPS FastRNN network uses highly optimized cells that avoid exploding and vanishing gradient problems associated with other RNN cells. Given that the FastRNN cells have shown state-of-the-art accuracy results with lesser parameters than other RNN Cell types, they represent a great benchmark to identify whether a compression technique is effective. As shown in Table \ref{tab:kprnnnresults}, we are able to compress these highly optimized cells using KPs by a factor of $16\times$ with minimal loss in accuracy, unlike alternative techniques. The KWS GRU network is the only entry which does not show results for magnitude pruning. This is because the magnitude pruning infrastructure we used \cite{suyog} is not available for GRU-based networks. The KP-based network is still more accurate than the remaining alternatives.

Additional details about how these experiments were run, the mean and variance of the accuracy, etc. can be found in Appendix \ref{sec:appkpkwslstm} and \ref{sec:appkpkwsgru}.

\label{sec:kpquant}
\begin{table}
\centering
\caption{Accuracy of baseline HAR1, baseline KWS-LSTM, KP compressed HAR1-BiLSTM and KP compressed KWS-LSTM network after quantization to 8-bits.}
\label{tab:quant}
\begin{tabular}{|l|l|l|l|l|}
\hline
 & \multicolumn{4}{c|}{HAR1} \\ \hline
 & \multicolumn{2}{c|}{32-bit} & \multicolumn{2}{c|}{8-bit} \\ \hline
 & Accuracy & \begin{tabular}[c]{@{}c@{}}Size \\ (KB)\end{tabular} & Accuracy & \begin{tabular}[c]{@{}c@{}}Size \\ (KB)\end{tabular} \\ \hline
Baseline & 91.90 & 1462.84 & 91.13 & 384.64 \\ \hline
KP & 91.14 & 74.91 & 90.90 & 28.22 \\ \hline
 & \multicolumn{4}{c|}{KWS-LSTM} \\ \hline
 & \multicolumn{2}{c|}{32-bit} & \multicolumn{2}{c|}{8-bit} \\ \hline
 & Accuracy & \begin{tabular}[c]{@{}c@{}}Size \\ (KB)\end{tabular} & Accuracy & \begin{tabular}[c]{@{}c@{}}Size \\ (KB)\end{tabular} \\ \hline
Baseline & 92.50 & 243.42 & 92.02 & 65.04 \\ \hline
KP & 91.20 & 15.30 & 91.04 & 8.01 \\ \hline
\end{tabular}
\end{table}

\begin{table*}[htb]
\centering
\begin{threeparttable}
\caption{Model accuracy and runtime for 2 different benchmarks before and after compression. This table shows that HKP can control the amount of compression in a KPRNN network. The baseline networks are compared to baseline networks with RNN layers compressed by either the KP, magnitude pruning, LMF, HKP or scaling the network size (Small Baseline). Each compressed network has fewer RNN parameters than the baseline (compression factor indicated). For HAR1-BiLSTM network, KP compressed the baseline by 28x. Using HKP, we were able to decrease the compression factor of the baseline to 20x and 10x and regain accuracy. Similarly, for KWS-LSTM, KP compressed the baseline network by 25x. Using HKP, we were able to decrease the compression factor of the baseline and reduce it to 20x and 10x to regain accuracy.}
\label{tab:hkprnnresults}
\begin{tabular}{|c|c|c|c|c|c|c|c|c|c|}
\hline
\multicolumn{1}{|c|}{} & \begin{tabular}[c]{@{}c@{}}Compression\\ Factor\end{tabular} & Network & \begin{tabular}[c]{@{}c@{}}Accuracy \\ (\%)\end{tabular} & \begin{tabular}[c]{@{}c@{}}Runtime \\ (ms)\end{tabular} & \multicolumn{1}{c|}{} & \begin{tabular}[c]{@{}c@{}}Compression\\ Factor\end{tabular} & Network & \begin{tabular}[c]{@{}c@{}}Accuracy \\ (\%)\end{tabular} & \begin{tabular}[c]{@{}c@{}}Runtime \\ (ms)\end{tabular} \\ \hline
\multirow{13}{*}{\begin{tabular}[c]{@{}c@{}}HAR1-\\ BiLSTM\end{tabular}} & $1\times$ & \textit{Baseline} & \textit{91.9} & \textit{470} & \multirow{13}{*}{\begin{tabular}[c]{@{}c@{}}KWS-\\ LSTM\end{tabular}} & $1\times$ & \textit{Baseline} & \textit{92.5} & \textit{26.8} \\ \cline{2-5} \cline{7-10} 
 & \multirow{4}{*}{$28\times$} & SB & 88.84 & 29.92 &  & \multirow{4}{*}{$25\times$} & SB & 89.7 & 2.01 \\ \cline{3-5} \cline{8-10} 
 &  & P & 82.97 & 98.2 &  &  & P & 84.91 & 5.9 \\ \cline{3-5} \cline{8-10} 
 &  & LMF & 89.94 & 64.12 &  &  & LMF & 89.13 & 4.1 \\ \cline{3-5} \cline{8-10} 
 &  & KP\tnote{1} & \textbf{91.14} & 157 &  &  & KP\tnote{1} & \textbf{91.2} & 17.5 \\ \cline{2-5} \cline{7-10} 
 & \multirow{4}{*}{$20\times$} & SB & 89.14 & 35.25 &  & \multirow{4}{*}{$20\times$} & SB & 89.8 & 2.25 \\ \cline{3-5} \cline{8-10} 
 &  & P & 86.7 & 103.31 &  &  & P & 85.17 & 8 \\ \cline{3-5} \cline{8-10} 
 &  & LMF & 90.6 & 72.60 &  &  & LMF & 90.9 & 5.78 \\ \cline{3-5} \cline{8-10} 
 &  & HKP\tnote{2} & \textbf{91.25} & 306.90 &  &  & HKP\tnote{2} & \textbf{91.28} & 15 \\ \cline{2-5} \cline{7-10} 
 & \multirow{4}{*}{$10\times$} & SB & 90.30 & 63.42 &  & \multirow{4}{*}{$10\times$} & SB & 89.70 & 3.2 \\ \cline{3-5} \cline{8-10} 
 &  & P & 89.20 & 174.92 &  &  & P & 89.49 & 11.26 \\ \cline{3-5} \cline{8-10} 
 &  & LMF & 90.80 & 87.94 &  &  & LMF & 91.50 & 6.9 \\ \cline{3-5} \cline{8-10} 
 &  & HKP\tnote{2} & \textbf{91.64} & 234.67 &  &  & HKP\tnote{2} & \textbf{91.95} & 15 \\ \hline
\end{tabular}
\begin{tablenotes}
    \item[1] KP calculates the point of maximum compression
    \item[2] HKP is used to control the amount of compression for a KPRNN network. Using HKP we are able to reduce the amount of compression to $20\times$ and $10\times$
\end{tablenotes}
\end{threeparttable}
\vspace{-1.0em}
\end{table*}

\subsubsection{Quantization} 

Quantization \cite{Quant1,QuantFlow,Quant2,Quant3,Quant4,Quant5,Quant-hubara,Quant-bengio} is another popular technique for compressing neural networks. Quantization is orthogonal to the compression techniques discussed previously. Prior work has shown that pruning~\cite{han2015deep_compression} can benefit from quantization. We conducted a study to test whether KPRNNs are compatible with quantization using the approach in \cite{QuantFlow}. We quantized the LSTM cells in the baseline and the KPRNN compressed networks to 8-bits floating point representations to test the robustness of KPRNNs under reduced bit-precision. Table~\ref{tab:quant} shows that quantization works well with KPRNNs. Quantization leads to an overall compression factor of $50\times$ and $30\times$ of the LSTM layers of HAR1-BiLSTM and KWS-LSTM networks and accuracy losses of $0.24\%$ and $0.16\%$, respectively. These results indicate that KP compression is compatible with quantization.

\subsection{Possible explanation for the accuracy difference between KPRNN, pruning, and LMF} 
In general, the poor accuracy of LMF can be attributed to significant reduction in the rank of the matrix (generally $<10$). KPs, on the other hand, will create a full rank matrix if the Kronecker factors are fully ranked. This is because of the following relationship \cite{laub2005matrix} -
\begin{gather}
    \mathrm{rank} \left(\;A \otimes B\right) \,=\, \mathrm{rank}\;A \cdot \mathrm{rank} \; B.
\end{gather}
We observe that, Kronecker factors of all the compressed benchmarks are fully-ranked. A full-rank matrix can also lead to poor accuracy if it is ill-conditioned \cite{Goodfellow}. However, KPRNN learns matrices that do not exhibit this behavior. The condition numbers of the matrices of the best-performing KP compressed networks discussed in this paper are in the range of $1.2$ to $7.3$.

To prune a network to the same compression factor as KP, networks need to be pruned to 94\% sparsity or above. It has been observed that pruning leads to an accuracy drop beyond 90\% sparsity \cite{bestprune}. Pruning FastRNN cells to the required compression factor leads to an ill-conditioned matrix. This may explain the poor accuracy of sparse FastRNN networks. However, for other pruned networks, the resultant sparse matrices have a condition number less than $20$ and are fully-ranked. Thus, condition number does not explain the loss in accuracy for these benchmarks. 

To further understand the loss in accuracy of pruned LSTM networks, we looked at the singular values of the resultant sparse matrices in the KWS-LSTM network. Let $y = Ax$. The largest singular value of $A$ upper-bounds $\Vert y \Vert_2$, i.e. the amplification applied by $A$. Thus, a matrix with larger singular value can lead to an output with larger norm \cite{linalgbook}. 
Since RNNs execute a matrix-vector product followed by a non-linear sigmoid or tanh layer, the output will saturate if the value is large. The matrix in the LSTM layer of the best-performing pruned KWS-LSTM network has its largest singular value in the range of $48$ to $52$ while the baseline KWS-LSTM network learns a LSTM layer matrix with largest singular value of $19$ and the Kronecker product compressed KWS-LSTM network learns LSTM layers with singular values less than $15$. This might explain the especially poor results achieved after pruning this benchmark. Similar observations can be made for the pruned HAR1 network.

\begin{table*}[tb]
\centering
\caption{Ablation study to compare HKP with Method 1 (Section \ref{sec:controlkp}). Method 1 is KP at non-maximum points of compression (KPNM). We compress the RNN layers in KWS-LSTM and HAR1-BiLSTM networks to 20x and 10x compression factors using HKP and KPNM. We observe that HKP generally provides an accuracy advantage over KPNM.}
\label{tab:ablation}
\begin{tabular}{|c|c|c|c|c|c|c|c|}
\hline
 & \begin{tabular}[c]{@{}c@{}}Compression\\ Factor\end{tabular} & Network & \begin{tabular}[c]{@{}c@{}}Accuracy\\ (\%)\end{tabular} &  & \begin{tabular}[c]{@{}c@{}}Compression\\ Factor\end{tabular} & Network & \begin{tabular}[c]{@{}c@{}}Accuracy\\ (\%)\end{tabular} \\ \hline
\multirow{5}{*}{\begin{tabular}[c]{@{}c@{}}HAR1-\\ BiLSTM\end{tabular}} & 1x & Baseline & 91.9 & \multirow{5}{*}{\begin{tabular}[c]{@{}c@{}}KWS-\\ LSTM\end{tabular}} & 1x & Baseline & 92.5 \\ \cline{2-4} \cline{6-8} 
 & \multirow{2}{*}{20x} & HKP & 91.25 &  & \multirow{2}{*}{20x} & HKP & 91.28 \\ \cline{3-4} \cline{7-8} 
 &  & KPNM & 90.40 &  &  & KPNM & 90.95 \\ \cline{2-4} \cline{6-8} 
 & \multirow{2}{*}{10x} & HKP & 91.64 &  & \multirow{2}{*}{10x} & HKP & 91.95 \\ \cline{3-4} \cline{7-8} 
 &  & KPNM & 91.17 &  &  & KPNM & 92.17 \\ \hline
\end{tabular}
\end{table*}

\subsection{HKPRNN Networks}

As mentioned in Section \ref{sec:hkprnn}, we target the point of maximum compression using the two-matrix Kronecker Product technique and using the HKPRNN technique is a useful way to control the level of compression and the corresponding reduction in accuracy and runtime.

Table \ref{tab:hkprnnresults} shows the results from using HKPRNN. This is a similar table as in Table \ref{tab:kprnnnresults}, but rather than using the only compression factor allowed by KPRNN, three possible compression ratios were explored -- $10\times$, $20\times$, and the maximum compression ratio --  resulting in the three data points for each compression scheme. The maximum compression ratio for the Kronecker Product technique is when a hybrid scheme is not used at all (i.e., $r=0$), so the HKPRNN data points at the maximum compression ratio are equivalent to the corresponding KPRNN data points in Table \ref{tab:kprnnnresults}. The results clearly indicate that using HKPRNN we are able to trade-in parameters to regain accuracy.


Additional details about the training hyperparameters used, the mean and variance of the accuracy and the specific model sizes and run-times can be found in Appendix \ref{sec:apphkphar1} and \ref{sec:apphkpkwslstm}.

\iftrue
\subsubsection{Ablation study to compare HKP with KP at non-maximum points of compression}
\label{sec:ablation}

As mentioned in Section \ref{sec:controlkp}, there are other methods apart from HKP to inject more parameters into KP compressed networks. However, HKP had specific practical advantages over each one of them. Specifically, HKP provides more points of compression than Method 1. We did an ablation study to understand if HKP had accuracy advantages over Method 1. We compressed the KWS-LSTM and HAR1-BiLSTM networks to the same compression points using Method 1 and HKP. The results are show in Table \ref{tab:ablation} and they indicate that HKP has accuracy advantage over Method 1 apart from the practical advantages. 

\fi

%% file: MainPaper/Conclusion.tex
\section{Conclusion}
\label{sec:conclusion}
We show how to compress RNN Cells by $16\times$ to $38\times$ using Kronecker products. We call the cells compressed using Kronecker products as KPRNNs. KPRNNs can act as a drop in replacement for most RNN layers and provide the benefit of significant compression with marginal impact on accuracy. Additionally, we show how to recover the lost accuracy by injecting small number of parameters into the KP compressed network. We call this family of controlled Kronecker compressed network as HKPRNN and show how we can compress the network by a factor of $10-20\times$. None of the other compression techniques (pruning, LMF) match the accuracy of the Kronecker compressed networks. We show that this compression technique works across 5 benchmarks that represent key applications in the IoT domain.

%% file: Appendix/Proof.tex
\section{Proof of the Matrix-Vector Multiplication Algorithm when the Matrix is expressed as a Kronecker product of two matrices}
\label{app:kpproofmv}
Let,
\begin{equation}
    y = (A\otimes B)\times x
\end{equation}
where, $y\in \mathbb{R}^{m\times 1}$, $x\in \mathbb{R}^{n\times 1}$, $A\in \mathbb{R}^{m1\times n1}$, $B\in \mathbb{R}^{m2\times n2}$ and $m=m1\times m2$, $n=n1\times n2$. 

\[
x = \left( \begin{array}{c}
x_{1} \\
x_{2} \\
x_{3} \\
. \\
. \\
. \\
x_{n1}
\end{array} \right)
y = \left( \begin{array}{c}
y_{1} \\
y_{2} \\
y_{3} \\
. \\
. \\
. \\
y_{m1}
\end{array} \right)
\]
where $x_{i} \in \mathbb{R}^{n2}$ and $y_{i} \in \mathbb{R}^{m2}$
Then, 
\begin{equation}
    y = (A\otimes B)\times x
\end{equation}

\[
y=
\left[ {\begin{array}{cccc}
a\textsubscript{1,1}B & a\textsubscript{1,2} B & ... & a\textsubscript{1,n1}B\\
a\textsubscript{2,1}B & a\textsubscript{2,2} B & ... & a\textsubscript{2,n1}B\\
. & . & . & . \\
. & . & . & . \\
a\textsubscript{m1,1} B & a\textsubscript{1,2} B & ... & a\textsubscript{m1,n1} B\\
\end{array} } \right]
\left[ \begin{array}{c}
x_{1} \\
x_{2} \\
x_{3} \\
. \\
. \\
. \\
x_{n1}
\end{array} \right]
\]

\[
\left[ \begin{array}{c}
y_{1} \\
y_{2} \\
y_{3} \\
. \\
. \\
. \\
y_{m1}
\end{array} \right]
= \left[ {\begin{array}{c}
a\textsubscript{1,1}Bx_{1} + a\textsubscript{1,2}Bx_{2} + ... + a\textsubscript{1,n1}Bx_{n1}\\
a\textsubscript{2,1}Bx_{1} + a\textsubscript{2,2}Bx_{2} + ... + a\textsubscript{2,n1}Bx_{n1}\\
. \\
. \\
a\textsubscript{m1,1}Bx_{1} + a\textsubscript{m1,2}Bx_{2} + ... + a\textsubscript{m1,n1}Bx_{n1}\\
\end{array} } \right]
\]

Each $y_{i}$ has the following form - 
\[
\left[ \begin{array}{c}
a\textsubscript{i,1}Bx_{1} + a\textsubscript{i,2}Bx_{2} + ... + a\textsubscript{i,n1}Bx_{n1}\\
\end{array} \right]
\]

\[
 = B\left[ \begin{array}{cccccc}
x_{1}  & x_{2} & . & . & . & x_{n1}\\
\end{array} \right]
\left[ \begin{array}{c}
a_{i,1}\\
a_{i,2}\\
. \\
. \\
a_{i,n1}\\
\end{array} \right]
\]

Now, let 
\[
X = \left[ \begin{array}{ccccc}
x_{1} & x_{2} & . & . & x_{n1}\\
\end{array} \right]
\]

And,
\[
\bm{a_{i}} = \left[ \begin{array}{ccccc}
a_{i,1} & a_{i,2} & . & . & a_{i,n1}\\
\end{array} \right]^T
\]
Then,
\begin{equation}
    y\textsubscript{i} = BX\textbf{a\textsubscript{i}}  \;(for \; i=1,2,...m1)
\end{equation}

Let \textbf{Y} be a concatenation of y\textsubscript{i}
Thus, 
\[
\bm{Y} = \left[ \begin{array}{ccccc}
y_{i} & y_{i} & . &. & y_{m1}\\
\end{array} \right]
\]
\[
\bm{Y} = \left[ \begin{array}{ccccc}
BX\bm{a_{1}} & BX\bm{a_{2}} & . &. & BX\bm{a_{m1}}\\
\end{array} \right]
\]
\[
\bm{Y} = BX\left[ \begin{array}{ccccc}
\bm{a_{1}} & \bm{a_{2}} & . &. & \bm{a_{m1}}\\
\end{array} \right]
\]
\[
\bm{Y} =BXA^T
\]

%% file: Appendix/Dataset.tex
\section{Dataset details and baseline implementation}
\label{sec:appDataset}

\subsection{Datasets}
\label{sec:appDatasetDeets}
We evaluate the impact of compression using the techniques discussed in section \ref{sec:kprnn} and \ref{sec:hkprnn} on a wide variety of benchmarks spanning applications like key-word spotting, human activity recognition, image classification and language modeling.
\begin{itemize}
    \item \textbf{Human Activity Recognition}: We use the \cite{opp} dataset for human activity recognition. We split the benchmark into training, validation and test dataset using the procedure described in \cite{hammerla2016deep}. They use a subset of 77 sensors from the dataset. They use run 2 from subject 1 as their validation set, and replicate the most popular recognition challenge by using runs 4 and 5 from subject 2 and 3 in the test set. The remaining data is used for training. For frame-by-frame analysis, they created sliding windows of duration 1 second and 50\% overlap leading to input vector of size 81x77 i.e. 81 dimensional input is fed to the network over 77 time steps. The resulting training-set contains approx. 650k samples (43k frames). 
    \item \textbf{Image Classification}: We use the MNIST \cite{mnist} and USPS \cite{usps} dataset for image classification. The USPS dataset consists of 7291 train and 2007 test images while the MNIST dataset consists of 60k training and 10k test images. We split the publicly available training set into 80\% training set and 20\% validation set and use the selected set of hyperparameters on the test set.
    \item \textbf{Key-word Spotting}: We use the \cite{warden} dataset for key-word spotting. The entire dataset consists of 65K different samples of 1-second long audio clips of 30 keywords, collected from thousands of people. We split the benchmark into training, validation and test dataset using the procedure described in \cite{zhang2017}.
\end{itemize}

\subsection{Data Pre-processing}
\label{sec:appDataPreProcess}
For the key-word spotting benchmarks, we reuse the framework provided by \cite{zhang2017}. Thus we pre-process the data as suggested by them. For the human activity recognition dataset, we follow the pre-processing procedure described in \cite{hammerla2016deep}. We reuse the framework provided by \cite{msr} for the USPS dataset, thus using the pre-processing procedure provided by them. 

\subsection{Baseline Algorithms and Implementation}
\label{sec:appBaselineAlgorithmsAndImplementation}
\begin{itemize}
    \item \textbf{MNIST}: For this benchmark, the $28\times 28$ image is fed to a single layer LSTM network with hidden vector of size 40 over 28 time steps. The dataset is fed using a batch size of 128 and the model is trained for 3000 epochs using a learning rate of 0.001. We use the Adam Optimizer \cite{adam} during training. Additionally, we divide the learning rate by 10 after every 1000 epochs. The total size of the network is 44.72 KB. 
    \item \textbf{HAR1}: We use the network described in \cite{hammerla2016deep}. Their network uses a bidirectional LSTM with hidden length of size 179 followed by a softmax layer to get an accuracy of 92.5\%. Input is of dimension 77 and is fed over 81 time steps. The paper uses gradient clipping regularization with a max norm value of 2.3 and a dropout of value 0.92 for both directions of the LSTM network. The network is trained for 300 epochs using a learning rate of 0.025, Adam optimization \cite{adam} and a batch size of 64. We used their training infrastructure and recreated the network in tensorflow. The suggested hyperparameters in the paper got us an accuracy of 91.9\%. Even after significant effort, we were not able get to the accuracy mentioned in the paper. Henceforth, we will use 91.9\% as the baseline accuracy. The total size of the network is 1462.836 KB.
    \item \textbf{KWS-LSTM}: For our baseline Basic LSTM network, we use the smallest LSTM model in \cite{zhang2017}. The input to the network is 10 MFCC features fed over 25 time steps. The LSTM architecture uses a hidden length of size 118 and achieves an accuracy of 92.50\%. We use a learning rate of 0.0005, 0.0001, 0.00002 for 10000 steps each with ADAM optimizer \cite{adam} and a batch size of 100. The total size of the network is 243.42 KB.
    \item \textbf{KWS-GRU}: We use the smallest GRU model in \cite{zhang2017} as our baseline. The input to the network is 10 MFCC features fed over 25 time steps. The GRU architecture uses a hidden length of size 154 and achieves an accuracy of 93.50\%. We use a learning rate of 0.0005,0.0001,0.00002 for 10000 steps each with ADAM optimizer and a batch size of 100. The total size of the network is 305.03 KB.
    \item \textbf{USPS-FastRNN}: The input image of size $16\times 16$ is divided into rows of size 16 that is fed into a single layer of FastRNN network \cite{msr} with hidden vector of size 32 over 16 time steps. The network is trained for 300 epochs using an initial learning rate of 0.01 and a batch size of 100. The learning starts rate declining by 0.1 after 200 epochs. The total size of the network is 7.54 KB. 
\end{itemize}

%% file: Appendix/KroneckerProduct.tex
\section{Kronecker Products - Implementation}
\label{sec:appKP}

\begin{algorithm*}
   \caption{Implementation of Kronecker Products in Tensorflow}
   \label{alg:kptf}
   \textbf{Input}: Matrices $B$ of dimension $m1 \times n1$, $C$ of dimension $m2 \times n2$ \\
   \textbf{Output}: Matrix $A$ of dimension $m\times n$ 
   \begin{algorithmic}[1]
   \STATE $b\_shape = [B.shape[0].value,B.shape[1].value]$
   \STATE $c\_shape = [C.shape[0].value,C.shape[1].value]$
   \STATE $temp1 = tf.reshape(B,[b\_shape[0],1,b\_shape[1],1])$
   \STATE $temp2 = tf.reshape(C,[1,c\_shape[0],1,c\_shape[1]])$
   \STATE $A = tf.reshape(temp1*temp2,[b\_shape[0]*c\_shape[0],b\_shape[1]*c\_shape[1]])$
\end{algorithmic}
\end{algorithm*}

%% file: Appendix/KPRNNHyperparameters.tex
\section{KPRNN - Additional Details}
\label{sec:appKPRNN}
\subsection{Hyperparameters}
\label{sec:appKPRNN_hp}
\begin{algorithm}[htbp]
   \caption{LRD1: Learning rate decay function}
   \label{alg:lrd1}
   \textbf{Input}: curr\_learning\_rate, decay\_rate, global\_step, decay\_steps\\
   \textbf{Output}: new\_learning\_rate 
   \begin{algorithmic}[1]
   \STATE $temp1 = global\_step/decay\_steps$
   \STATE $pow = decay\_rate^{temp1}$
   \STATE $new\_learning\_rate = curr\_learning\_rate*pow$
\end{algorithmic}
\end{algorithm}

\subsubsection{MNIST-LSTM compressed using KPLSTM Cells}
\label{sec:appkpmnist}
\textbf{Hyperparameters}: Table \ref{tab:hp-kpmnist} shows the hyperparameters used for training the MNIST-LSTM baseline and the MNIST network compressed using pruning, LMF, KPLSTM and a smaller baseline with the number of parameters equivalent to the compressed network. 

\textbf{Mean and Std Deviation of the accuracy of the compressed network}: Last three rows of Table \ref{tab:hp-kpmnist} show the top test accuracy, mean test accuracy and standard deviation of test accuracy of the networks trained using top two sets of best performing hyper-parameters on a held out validation set. 

\textbf{Hyperparameter values explored}: We explored a broad range of hyper-parameter that were the intersection of the following values -
\begin{itemize}
    \item Initial Learning Rate - 0.01 to 0.001 in multiples of 3
    \item LR Decay Schedule - We experimented with a step function and exponential decay function as described in algorithm \ref{alg:lrd1}.
\end{itemize}

\subsubsection{HAR1 compressed using KPLSTM Cells}
\label{sec:appkphar1}
\textbf{Hyperparameters}: Table \ref{tab:hp-kphar1} shows the hyperparameters used for training the HAR1 baseline and the HAR1 network compressed using pruning, LMF, KPLSTM and a smaller baseline with the number of parameters equivalent to the compressed network. 

\textbf{Mean and Std Deviation of the accuracy of the compressed network}: Last three rows of Table \ref{tab:hp-kphar1} show the top test accuracy, mean test accuracy and standard deviation of test accuracy of the networks trained using top two sets of best performing hyper-parameters on a held out validation set. 

\textbf{Hyperparameter values explored}: We explored a broad range of hyper-parameter that were the intersection of the following values -
\begin{itemize}
    \item Initial Learning Rate - 0.0025 to 0.25 in multiples of 3
    \item Max Norm - 1, 1.5, 2.3 and 3.5
    \item Dropout - 0.3, 0.5, 0.7 and 0.9
    \item \#Epochs - 200 to 400 in increments of 100 for all networks apart from pruning. For pruned networks, we increased the number of epochs to 600
    \item LR Decay Schedule - We experimented with a step function and exponential decay function as described in algorithm \ref{alg:lrd1}.
    \item Pruning parameters - We explored various pruning start\_epoch and end\_epoch. We looked at starting pruning after 25\% to 33\% of the total epochs in increments of 4\% and ending pruning at 75\% to 83\% of the total epochs in increments of 4\%
\end{itemize}

\subsubsection{KWS-LSTM compressed using KPLSTM Cells}
\label{sec:appkpkwslstm}
\textbf{Hyperparameters}: Table \ref{tab:hp-kpkwslstm} shows the hyperparameters used for training the HAR1 baseline and the HAR1 network compressed using pruning, LMF, KPLSTM and a smaller baseline with the number of parameters equivalent to the compressed network. 

\textbf{Mean and Std Deviation of the accuracy of the compressed network}: Last three rows of Table \ref{tab:hp-kpkwslstm} show the top test accuracy, mean test accuracy and standard deviation of test accuracy of the networks trained using top two sets of best performing hyper-parameters on a held out validation set. 

\textbf{Hyperparameter values explored}: We explored a broad range of hyper-parameter that were the intersection of the following values -
\begin{itemize}
    \item Initial Learning Rate - 0.001 to 0.1 in multiples of 10
    \item \#Epochs - We trained the network for 30k-100k epochs with increments of 10k
    \item LR Decay Schedule - We experimented with a step function and exponential decay function as described in algorithm \ref{alg:lrd1}. For the step function we decremented the learning rate by 10 after every 10k, 20k or 30k steps depending on the improvement in held out validation accuracy. For the LRD1 algorithm, we tried decay\_rate values of 0.03 to 0.09 in increments of 0.02.
    \item Pruning parameters - We explored various pruning start\_epoch and end\_epoch. We looked at starting pruning after 10k to 25k in increments of 5k and ending pruning at 60k to 90k in increments of 10k
\end{itemize}

\subsubsection{KWS-GRU compressed using KPGRU Cells}
\label{sec:appkpkwsgru}
\textbf{Hyperparameters}: Table \ref{tab:hp-kpkwsgru} shows the hyperparameters used for training the HAR1 baseline and the HAR1 network compressed using pruning, LMF, KPGRU and a smaller baseline with the number of parameters equivalent to the compressed network. 

\textbf{Mean and Std Deviation of the accuracy of the compressed network}: Last three rows of Table \ref{tab:hp-kpkwsgru} show the top test accuracy, mean test accuracy and standard deviation of test accuracy of the networks trained using top two sets of best performing hyper-parameters on a held out validation set. 

\textbf{Hyperparameter values explored}: We explored a broad range of hyper-parameter that were the intersection of the following values -
\begin{itemize}
    \item Initial Learning Rate - 0.001 to 0.1 in multiples of 10
    \item \#Epochs - We trained the network for 30k-100k epochs with increments of 10k
    \item LR Decay Schedule - We experimented with a step function and exponential decay function as described in algorithm \ref{alg:lrd1}. For the step function we decremented the learning rate by 10 after every 10k, 20k or 30k steps depending on the improvement in held out validation accuracy. For the LRD1 algorithm, we tried decay\_rate values of 0.03 to 0.09 in increments of 0.02.
    \item Pruning parameters - We explored various pruning start\_epoch and end\_epoch. We looked at starting pruning after 10k to 25k in increments of 5k and ending pruning at 60k to 90k in increments of 10k
\end{itemize}

\subsubsection{USPS-FastRNN compressed using KPFastRNN Cells}
\label{sec:appkpkwsusps}
\textbf{Hyperparameters}: Table \ref{tab:hp-uspskp} shows the hyperparameters used for training the HAR1 baseline and the HAR1 network compressed using pruning, LMF, KPFastRNN and a smaller baseline with the number of parameters equivalent to the compressed network. 

\textbf{Mean and Std Deviation of the accuracy of the compressed network}: Last three rows of Table \ref{tab:hp-uspskp} show the top test accuracy, mean test accuracy and standard deviation of test accuracy of the networks trained using top two sets of best performing hyper-parameters on a held out validation set. 

\textbf{Hyperparameter values explored}: We explored a broad range of hyper-parameter that were the intersection of the following values -
\begin{itemize}
    \item Initial Learning Rate - 0.01 to 0.001 in multiples of 3
    \item LR Decay Schedule - We experimented with a step function and exponential decay function as described in algorithm \ref{alg:lrd1}.
\end{itemize}

\begin{table*}
\centering
\caption{Hyperparameters for MNIST baseline network, network compressed using KPLSTM and equivalent sized networks compressed using LMF, Pruning and Small Baseline. LRD1 refers to Algorithm \ref{alg:lrd1}.}
\label{tab:hp-kpmnist}
\begin{tabular}{|c|c|c|c|c|c|}
\hline
Network & Baseline & \begin{tabular}[c]{@{}c@{}}Small\\ Baseline\end{tabular} & Pruning & LMF & KPLSTM \\ \hline
\begin{tabular}[c]{@{}c@{}}Batch\\ Size\end{tabular} & 128 & 128 & 128 & 128 & 128 \\ \hline
Optimizer & Adam & Adam & Adam & Adam & Adam \\ \hline
\begin{tabular}[c]{@{}c@{}}Weight\\  Init\end{tabular}  &\multicolumn{5}{c|}{glorot\_uniform} \\ \hline
\#Epochs & 3000 & 4000 & 10000 & 5000 & 5000 \\ \hline
\begin{tabular}[c]{@{}c@{}}Initial\\ LR\end{tabular} & 0.001 & 0.001 & 0.001 & 0.001 & 0.001 \\ \hline
\begin{tabular}[c]{@{}c@{}}Decay\\ Schedule\end{tabular} & \multicolumn{5}{c|}{LR is divided by 10 after every \#Epochs$\div 4$ epochs} \\ \hline
Additional Details &  &  &  &  &  \\ \hline
\#Layers & 1 & 1 & 1 & 1 & 1 \\ \hline
Hidden Vector Size & 40 & 40 & 40 & 40 & 40 \\ \hline
Size of input & 28 & 28 & 28 & 28 & 28 \\ \hline
\#Time Steps & 28 & 28 & 28 & 28 & 28 \\ \hline
\begin{tabular}[c]{@{}c@{}}Size (KB)\\ for 32 bit weights\end{tabular} & 44.73 & 4.51 & 4.19 & 4.9 & 4.05 \\ \hline
\begin{tabular}[c]{@{}c@{}}Mean\\Accuracy\end{tabular} & - & 87.20 & 96.49 & 97.24 & 98.28 \\ \hline
\begin{tabular}[c]{@{}c@{}}Top\\Accuracy\end{tabular} & 99.40 & 87.50 & 96.81 & 97.40 & 98.44 \\ \hline
\begin{tabular}[c]{@{}c@{}}Std\\Dev\\(Accuracy)\end{tabular} & - & 0.27 & 0.30 & 0.13 & 0.12 \\ \hline
\begin{tabular}[c]{@{}c@{}}Runtime\\(ms)\end{tabular} & 6.4 & 0.8 & 0.66 & 1.8 & 5.6 \\ \hline
\end{tabular}
\end{table*}

\begin{table*}
\centering
\caption{Hyperparameters for HAR1 baseline network, network compressed using KPLSTM and equivalent sized networks compressed using LMF, Pruning and Small Baseline. LRD1 refers to Algorithm \ref{alg:lrd1}.}
\label{tab:hp-kphar1}
\begin{tabular}{|c|c|c|c|c|c|}
\hline
Network & Baseline & \begin{tabular}[c]{@{}c@{}}Small\\ Baseline\end{tabular} & Pruning & LMF & KPLSTM \\
\hline
Optimizer & Adam & Adam & Adam & Adam & Adam \\ \hline
\begin{tabular}[c]{@{}c@{}}Batch\\Size\end{tabular} & 64 & 64 & 64 & 64 & 64 \\ \hline
\begin{tabular}[c]{@{}c@{}}Weight\\ Init\end{tabular} &\multicolumn{5}{c|}{glorot\_uniform} \\ \hline
\#Epochs & 300 & 300 & 600 & 300 & 300 \\ \hline
\begin{tabular}[c]{@{}c@{}}Initial\\   \\     LR\end{tabular} & 0.025 & 0.025 & 0.025 & 0.025 & 0.025 \\ \hline
\begin{tabular}[c]{@{}c@{}}Decay\\   \\     Schedule\end{tabular} & \multicolumn{5}{c|}{LR reduced by a factor of 10 after every 100 epochs} \\ \hline
MaxNorm & 2.3 & 2.3 & 2.3 & 3.5 & 2.3 \\ \hline
Dropout & 0.92 & 0.92 & 0.7 & 0.5 & 0.5 \\ \hline
\#Layers & 1 & 1 & 1 & 1 & 1 \\ \hline
\begin{tabular}[c]{@{}c@{}}Hidden\\Vector\\Size\end{tabular} & 179 & 179 & 179 & 179 & 178 \\ \hline
\begin{tabular}[c]{@{}c@{}}Size of\\Input\end{tabular} & 77 & 77 & 77 & 77 & 77 \\ \hline
\#Time Steps & 81 & 81 & 81 & 81 & 81 \\ \hline
\begin{tabular}[c]{@{}c@{}}Additional\\Details\end{tabular} &  &  & \begin{tabular}[c]{@{}c@{}}Pruning starts at\\Epoch \#100 and\\   ends at Epoch \#500\end{tabular} &  &  \\ \hline
\begin{tabular}[c]{@{}c@{}}Size (KB)\\ for \\32-bit weights\end{tabular} & 1462.84 & 75.90 & 75.55 & 76.40 & 74.91 \\ \hline
\begin{tabular}[c]{@{}c@{}}Mean\\Accuracy\end{tabular} & - & 88.39 & 89.63 & 89.63 & 90.95 \\ \hline
\begin{tabular}[c]{@{}c@{}}Top\\Accuracy\end{tabular} & 91.90 & 88.84 & 89.94 & 89.94 & 91.14 \\ \hline
\begin{tabular}[c]{@{}c@{}}Std\\Dev\\(Accuracy)\end{tabular} & - & 0.49 & 0.41 & 0.23 & 0.14 \\ \hline
\begin{tabular}[c]{@{}c@{}}Runtime\\(ms)\end{tabular} & 470 & 29.92 & 98.2 & 64.12 & 187 \\ \hline
\end{tabular}
\end{table*}

\begin{table*}
\centering
\caption{Hyperparameters for KWS-LSTM baseline network, network compressed using KPLSTM and equivalent sized networks compressed using LMF, Pruning and Small Baseline. LRD1 refers to Algorithm \ref{alg:lrd1}.}
\label{tab:hp-kpkwslstm}
\begin{tabular}{|c|c|c|c|c|c|}
\hline
Network & Baseline & \begin{tabular}[c]{@{}c@{}}Small\\ Baseline\end{tabular} & Pruning & LMF & KPLSTM \\ \hline
\begin{tabular}[c]{@{}c@{}}Batch\\Size\end{tabular} & 100 & 100 & 100 & 100 & 100 \\ \hline
Optimizer & Adam & Adam & Adam & Adam & Adam \\ \hline
\begin{tabular}[c]{@{}c@{}}Weight\\Init\end{tabular} & \multicolumn{5}{c|}{glorot\_uniform} \\ \hline
\#Epochs & 30k & 80k & 100k & 80k & 90k \\ \hline
\begin{tabular}[c]{@{}c@{}}Initial\\   \\     LR\end{tabular} & 5x10\textasciicircum{}-4 & 10\textasciicircum{}-2 & 10\textasciicircum{}-2 & 10\textasciicircum{}-2 & 10\textasciicircum{}-2 \\ \hline
\begin{tabular}[c]{@{}c@{}}Decay\\Schedule\end{tabular} & \begin{tabular}[c]{@{}c@{}}5x10\textasciicircum{}-4,\\ 1x10\textasciicircum{}-4,\\ 2x10\textasciicircum{}-5 for\\ 10k steps each\end{tabular} & \begin{tabular}[c]{@{}c@{}}LRD1 \\     with \\ decay\_rate\\     0.09\end{tabular} & \begin{tabular}[c]{@{}c@{}}LRD1 \\     with \\ decay\_rate\\     0.09\end{tabular} & \begin{tabular}[c]{@{}c@{}}10\textasciicircum{}-2, 10\textasciicircum{}-3, \\ 5x10\textasciicircum{}-4, 10\textasciicircum{}-4,\\   10\textasciicircum{}-4 for 10k, \\ 20k, 15k, 10k, \\ 15k and 10k \\ epochs each\end{tabular} & \begin{tabular}[c]{@{}c@{}}LRD1 \\     with \\ decay\_rate\\     0.09\end{tabular} \\ \hline
\#Layers & 1 & 1 & 1 & 1 & 1 \\ \hline
\begin{tabular}[c]{@{}c@{}}Hidden\\Vector\\Size\end{tabular} & 118 & 118 & 118 & 118 & 118 \\ \hline
\begin{tabular}[c]{@{}c@{}}Size of\\Input\end{tabular} & 10 & 10 & 10 & 10 & 10 \\ \hline
\#Time Steps & 25 & 25 & 25 & 25 & 25 \\ \hline
\begin{tabular}[c]{@{}c@{}}Additional\\Details\end{tabular} &  &  & \begin{tabular}[c]{@{}c@{}}Pruning starts \\ at Epoch \#15k \\ and ends \\ at Epoch \#80k\end{tabular} &  &  \\ \hline
\begin{tabular}[c]{@{}c@{}}Size (KB)\\ for \\32 bit weights\end{tabular} & 243.42 & 15.66 & 15.57 & 16.80 & 15.30 \\ \hline
\begin{tabular}[c]{@{}c@{}}Mean\\Accuracy\end{tabular} &-  & 88.57 & 82.51 & 88.94 & 91.12 \\ \hline
\begin{tabular}[c]{@{}c@{}}Top\\Accuracy\end{tabular} &92.50  &89.70  & 84.91 & 89.13 & 91.20 \\ \hline
\begin{tabular}[c]{@{}c@{}}Std\\Dev\\(Accuracy)\end{tabular} &-  & 0.67 & 2.72 & 0.16 & 0.07 \\ \hline
\begin{tabular}[c]{@{}c@{}}Runtime\\(ms)\end{tabular} & 26.8 & 2.01 & 5.89 & 4.14 & 17.5 \\ \hline
\end{tabular}
\end{table*}

\begin{table*}
\centering
\caption{Hyperparameters for KWS-GRU baseline network, network compressed using KPGRU and equivalent sized networks compressed using LMF, Pruning and Small Baseline. LRD1 refers to Algorithm \ref{alg:lrd1}.}
\label{tab:hp-kpkwsgru}
\begin{tabular}{|c|c|c|c|c|c|c|}
\hline
Network & Baseline & \begin{tabular}[c]{@{}c@{}}Small \\Baseline \\ (1L)\end{tabular} & \begin{tabular}[c]{@{}c@{}}Small \\Baseline \\ (2L)\end{tabular} & \begin{tabular}[c]{@{}c@{}}LMF \\ (1L)\end{tabular} & \begin{tabular}[c]{@{}c@{}}LMF \\ (2L)\end{tabular} & KPGRU \\ \hline
\begin{tabular}[c]{@{}c@{}}Batch\\ Size\end{tabular} & 100 & 100 & 100 & 100 & 100 & 100 \\ \hline
Optimizer & Adam & Adam & Adam & Adam & Adam & Adam \\ \hline
\begin{tabular}[c]{@{}c@{}}Weight\\ Init\end{tabular} & \multicolumn{6}{c|}{glorot\_uniform} \\ \hline
\#Epochs & 30k & 70k & 70k & 90k & 70k & 90k \\ \hline
\begin{tabular}[c]{@{}c@{}}Initial\\ LR\end{tabular} & 5x10\textasciicircum{}-3 & 10\textasciicircum{}-2 & 10\textasciicircum{}-2 & 10\textasciicircum{}-2 & 10\textasciicircum{}-2 & 0.01 \\ \hline
\begin{tabular}[c]{@{}c@{}}Decay\\ Schedule\end{tabular} & \begin{tabular}[c]{@{}c@{}}5x10\textasciicircum{}-3,\\ 10\textasciicircum{}-3,\\2x10\textasciicircum{}-4 \\ for 10k\\steps\\ each\end{tabular} & \begin{tabular}[c]{@{}c@{}}10\textasciicircum{}-2,\\ 5x10\textasciicircum{}-3,\\ 10\textasciicircum{}-3,\\ 2x10\textasciicircum{}-4\\ for 15k, \\15k,\\ 15k,15k \\and 10k \\ epochs \\ each\end{tabular} & \begin{tabular}[c]{@{}c@{}}10\textasciicircum{}-2,\\ 5x10\textasciicircum{}-3,\\ 10\textasciicircum{}-3,\\ 2x10\textasciicircum{}-4 \\ for 20k, 15k, \\ 10k, 15k\\ and 10k \\ epochs \\each\end{tabular} & \begin{tabular}[c]{@{}c@{}}10\textasciicircum{}-2,\\ 5x10\textasciicircum{}-3,\\ 10\textasciicircum{}-3,\\ 10\textasciicircum{}-4 \\ for 20k,\\ 30k,20k\\ and 20k \\ epochs\\ each\end{tabular} & \begin{tabular}[c]{@{}c@{}}10\textasciicircum{}-2,\\ 5x10\textasciicircum{}-3,\\ 10\textasciicircum{}-3, 10\textasciicircum{}-4 \\ for 20k, 15k, \\ 10k, 15k\\ and 10k \\ epochs each\end{tabular} & \begin{tabular}[c]{@{}c@{}}LRD1 \\ with \\ decay\_rate\\ 0.01\end{tabular} \\ \hline
\begin{tabular}[c]{@{}c@{}}Additional\\ Details\end{tabular} &  &  &  &  &  &  \\ \hline
\#Layers & 1 & 1 & 2 & 1 & 2 & 2 \\ \hline
\begin{tabular}[c]{@{}c@{}}Hidden\\ Vector\\ Size\end{tabular} & 154 &  154&  154&  154&  154& 154 \\ \hline
\begin{tabular}[c]{@{}c@{}}Size of\\   Input\end{tabular} & 10 & 10 & 10 & 10 & 10 & 10 \\ \hline
\begin{tabular}[c]{@{}c@{}}\#Time\\ Steps\end{tabular} & 25 & 25 & 25 & 25 & 25 & 25 \\ \hline
\begin{tabular}[c]{@{}c@{}}Size (KB)\\ for \\32 bit\\ weights\end{tabular} & 305.04 & 22.63 & 22.27 & 24.50 & 25.47 & 22.23 \\ \hline
\begin{tabular}[c]{@{}c@{}}Mean\\ Accuracy\end{tabular} & - & 85.76 & 82.71 & 90.39 & 87.10 & 92.03 \\ \hline
\begin{tabular}[c]{@{}c@{}}Top\\ Accuracy\end{tabular} & 93.50 & 86.40 & 84.53 & 90.88 & 87.70 & 92.30 \\ \hline
\begin{tabular}[c]{@{}c@{}}Std\\Deviation\\(Accuracy)\end{tabular} & - & 0.52 & 1.22 & 0.44 & 0.44 & 0.22 \\ \hline
\begin{tabular}[c]{@{}c@{}}Runtime\\(ms)\end{tabular} & 67 & 6 & 10.13 & 7.16 & 11.1 &34 \\ \hline
\end{tabular}
\end{table*}

\begin{table*}
\centering
\caption{Hyperparameters for USPS-FastRNN baseline network, network compressed using KPGRU and equivalent sized networks compressed using LMF, Pruning and Small Baseline. LRD1 refers to Algorithm \ref{alg:lrd1}.}
\label{tab:hp-uspskp}
\begin{tabular}{|c|c|c|c|c|c|}
\hline
Network & Baseline & Small Baseline & Pruning & LMF & KPFastRNN \\ \hline
\begin{tabular}[c]{@{}c@{}}Batch\\ Size\end{tabular} & 100 & 100 & 100 & 100 & 100 \\ \hline
Optimizer & Adam & Adam & Adam & Adam & Adam \\ \hline
\begin{tabular}[c]{@{}c@{}}Weight\\ Init\end{tabular} & \multicolumn{5}{c|}{random\_normal} \\ \hline
\#Epochs & 300 & 400 & 500 & 500 & 500 \\ \hline
\begin{tabular}[c]{@{}c@{}}Initial\\ LR\end{tabular} & 0.01 & 0.01 & 0.01 & 0.01 & 0.01 \\ \hline
\begin{tabular}[c]{@{}c@{}}Decay\\ Schedule\end{tabular} & \multicolumn{5}{c|}{Learning Rate declines by 10 after 200th epoch} \\ \hline
\#Layers & 1 & 1 & 1 & 1 & 1 \\ \hline
\begin{tabular}[c]{@{}c@{}}Hidden\\ Vector\\ Size\end{tabular} & 32 & 32 & 32 & 32 & 32 \\ \hline
\begin{tabular}[c]{@{}c@{}}Size of\\ Input\end{tabular} & 16 & 16 & 16 & 16 & 16 \\ \hline
\begin{tabular}[c]{@{}c@{}}\#Time\\ Steps\end{tabular} & 16 & 16 & 16 & 16 & 16 \\ \hline
\begin{tabular}[c]{@{}c@{}}Additional\\ Details\end{tabular} &  &  &  &  &  \\ \hline
\begin{tabular}[c]{@{}c@{}}Size (KB)\\ for \\32 bit weights\end{tabular} & 7.25 & 1.98 & 1.92 & 2.05 & 1.63 \\ \hline
\begin{tabular}[c]{@{}c@{}}Mean\\Accuracy\end{tabular} & - & 91.13 & 86.57 & 89.39 & 93.16 \\ \hline
\begin{tabular}[c]{@{}c@{}}Top\\Accuracy\end{tabular} & 92.50 & 91.23 & 88.52 & 89.56 & 93.20 \\ \hline
\begin{tabular}[c]{@{}c@{}}Std\\Dev\\(Accuracy)\end{tabular} & - & 0.07 & 1.52 & 0.14 & 0.03 \\ \hline
\begin{tabular}[c]{@{}c@{}}Runtime\\(ms)\end{tabular} & 1.175 & 0.4 & 0.375 & 0.283 & 0.6 \\ \hline

\end{tabular}
\end{table*}

\subsection{Quantization}

%% file: Appendix/HKPRNNHyperparameters.tex
\section{HKPRNN - Additional Details}
\label{sec:appHKPRNN}
\subsection{Hyperparameters}
\label{sec:appHKPRNN_hp}

\subsubsection{HAR1 compressed using HKPLSTM}
\label{sec:apphkphar1}
\textbf{Hyperparameters}: Table \ref{tab:hp-har1hkp} shows the hyperparameters used for training the HAR1 baseline and the HAR1 network compressed using pruning, LMF, HKPLSTM and a smaller baseline with the number of parameters equivalent to the compressed network. 

\textbf{Mean and Std Deviation of the accuracy of the compressed network}: Last three rows of Table \ref{tab:hp-har1hkp} show the top test accuracy, mean test accuracy and standard deviation of test accuracy of the networks trained using top two sets of best performing hyper-parameters on a held out validation set. 

\textbf{Hyperparameter values explored}: We explored a broad range of hyper-parameter that were the intersection of the following values -
\begin{itemize}
    \item Initial Learning Rate - 0.0025 to 0.25 in multiples of 3
    \item Max Norm - 1, 1.5, 2.3 and 3.5
    \item Dropout - 0.3, 0.5, 0.7 and 0.9
    \item \#Epochs - 200 to 400 in increments of 100 for all networks apart from pruning. For pruned networks, we increased the number of epochs to 600
    \item LR Decay Schedule - We experimented with a step function and exponential decay function as described in algorithm \ref{alg:lrd1}.
    \item Pruning parameters - We explored various pruning start\_epoch and end\_epoch. We looked at starting pruning after 25\% to 33\% of the total epochs in increments of 4\% and ending pruning at 75\% to 83\% of the total epochs in increments of 4\%
\end{itemize}

\subsubsection{KWS-LSTM compressed using HKPLSTM}
\label{sec:apphkpkwslstm}
\textbf{Hyperparameters}: Table \ref{tab:hp-kwslstmhkp20},\ref{tab:hp-kwslstmhkp10} shows the hyperparameters used for training the KWS-LSTM baseline and the KWS-LSTM network compressed using pruning, LMF, HKPLSTM and a smaller baseline with the number of parameters equivalent to the compressed network.

\textbf{Mean and Std Deviation of the accuracy of the compressed network}: Last three rows of Table \ref{tab:hp-kwslstmhkp10},\ref{tab:hp-kwslstmhkp20} show the top test accuracy, mean test accuracy and standard deviation of test accuracy of the networks trained using top two sets of best performing hyper-parameters on a held out validation set. 

\textbf{Hyperparameter values explored}:
We explored a broad range of hyper-parameter that were the intersection of the following values -
\begin{itemize}
    \item Initial Learning Rate - 0.001 to 0.1 in multiples of 10
    \item \#Epochs - We trained the network for 30k-100k epochs with increments of 10k
    \item LR Decay Schedule - We experimented with a step function and exponential decay function as described in algorithm \ref{alg:lrd1}. For the step function we decremented the learning rate by 10 after every 10k, 20k or 30k steps depending on the improvement in held out validation accuracy. For the LRD1 algorithm, we tried decay\_rate values of 0.03 to 0.09 in increments of 0.02.
    \item Pruning parameters - We explored various pruning start\_epoch and end\_epoch. We looked at starting pruning after 10k to 25k in increments of 5k and ending pruning at 60k to 90k in increments of 10k
\end{itemize}

\begin{table*}
\centering
\caption{Hyperparameters for HAR1 baseline network, network with LSTM layers compressed using HKPLSTM by a factor of 10 and equivalent sized networks compressed using LMF, Pruning and Small Baseline. LRD1 refers to Algorithm \ref{alg:lrd1}.}
\label{tab:hp-har1hkp}
\begin{tabular}{|c|c|c|c|c|c|}
\hline
Network & Baseline & Small Baseline & Pruning & LMF & HKPLSTM \\ \hline
\begin{tabular}[c]{@{}c@{}}Batch\\Size\end{tabular} & 64 & 64 & 64 & 64 & 64 \\ \hline
Optimizer & Adam & Adam & Adam & Adam & Adam \\ \hline
\begin{tabular}[c]{@{}c@{}}Weight\\Init\end{tabular} & \multicolumn{5}{c|}{glorot\_uniform} \\ \hline
\#Epochs & 300 & 200 & 300 & 300 & 300 \\ \hline
\begin{tabular}[c]{@{}c@{}}Initial\\ LR\end{tabular} & 0.025 & 0.025 & 0.025 & 0.025 & 0.025 \\ \hline
\begin{tabular}[c]{@{}c@{}}Decay\\ Schedule\end{tabular} & \multicolumn{5}{c|}{LR reduced by a factor of 10 after every 100 epochs} \\ \hline
MaxNorm & 2.3 & 2.3 & 3.5 & 2.3 & 2.3 \\ \hline
Dropout & 0.92 & 0.8 & 0.92 & 0.5 & 0.5 \\ \hline
\begin{tabular}[c]{@{}c@{}}\#Bidirectional\\ Layers\end{tabular} & 1 & 1 & 1 & 1 & 1 \\ \hline
\begin{tabular}[c]{@{}c@{}}Hidden Vector\\ Size\end{tabular} & 179 & 179 & 179 & 179 & 179 \\ \hline
\begin{tabular}[c]{@{}c@{}}Size of\\  Input\end{tabular} & 77 & 77 & 77 & 77 & 77 \\ \hline
\#Time Steps & 81 & 81 & 81 & 81 & 81 \\ \hline
\begin{tabular}[c]{@{}c@{}}Additional\\   Details\end{tabular} &  &  & \begin{tabular}[c]{@{}c@{}}Pruning starts at\\ Epoch \#100 and\\ ends at \\ Epoch \#250\end{tabular} &  &  \\ \hline
\begin{tabular}[c]{@{}c@{}}Size (KB)\\ assuming \\32 bit weights\end{tabular} & 1462.84 & 173.94 & 169 & 167.53 & 159.83 \\
\hline
\begin{tabular}[c]{@{}c@{}}Mean\\Accuracy\end{tabular} & - & 89.95 & 86.56 & 90.61 & 91.025 \\ \hline
\begin{tabular}[c]{@{}c@{}}Top\\Accuracy\end{tabular} & 91.90 & 90.30 & 87.20 & 90.80 & 91.20 \\ \hline
\begin{tabular}[c]{@{}c@{}}Std\\Dev\\(Accuracy)\end{tabular} & - & 0.22 & 0.34 & 0.17 & 0.14 \\ \hline
\begin{tabular}[c]{@{}c@{}}Runtime\\(ms)\end{tabular} & 470 & 63.42 & 174.92 & 87.94 & 234.67 \\ \hline
\end{tabular}
\end{table*}

\begin{table*}
\centering
\caption{Hyperparameters for KWS-LSTM baseline network, network with LSTM layers compressed using HKPLSTM by a factor of 10 and equivalent sized networks compressed using LMF, Pruning and Small Baseline. LRD1 refers to Algorithm \ref{alg:lrd1}.}
\label{tab:hp-kwslstmhkp10}
\begin{tabular}{|c|c|c|c|c|c|}
\hline
Network & Baseline & Small Baseline & Pruning & LMF & HKPLSTM \\ \hline
\begin{tabular}[c]{@{}c@{}}Batch\\Size\end{tabular} & 100 & 100 & 100 & 100 & 100 \\ \hline
Optimizer & Adam & Adam & Adam & Adam & Adam \\ \hline
\begin{tabular}[c]{@{}c@{}}Weight\\Init\end{tabular} & \multicolumn{5}{c|}{glorot\_uniform} \\ \hline
\#Epochs & 30k & 80k & 100k & 80k & 90k \\ \hline
\begin{tabular}[c]{@{}c@{}}Initial\\   \\     LR\end{tabular} & 5x10\textasciicircum{}-4 & 10\textasciicircum{}-2 & 10\textasciicircum{}-2 & 10\textasciicircum{}-2 & 10\textasciicircum{}-2 \\ \hline
\begin{tabular}[c]{@{}c@{}}Decay\\Schedule\end{tabular} & \begin{tabular}[c]{@{}c@{}}5x10\textasciicircum{}-4,\\ 1x10\textasciicircum{}-4,\\ 2x10\textasciicircum{}-5 for\\ 10k steps each\end{tabular} & \begin{tabular}[c]{@{}c@{}}LRD1 \\     with \\ decay\_rate\\     0.09\end{tabular} & \begin{tabular}[c]{@{}c@{}}LRD1 \\     with \\ decay\_rate\\     0.09\end{tabular} & \begin{tabular}[c]{@{}c@{}}LRD1 \\     with \\ decay\_rate\\     0.09\end{tabular} & \begin{tabular}[c]{@{}c@{}}LRD1 \\     with \\ decay\_rate\\     0.09\end{tabular} \\ \hline
\#Layers & 1 & 1 & 1 & 1 & 1 \\ \hline
\begin{tabular}[c]{@{}c@{}}Hidden\\Vector\\Size\end{tabular} & 118 & 118 & 118 & 118 & 118 \\ \hline
\begin{tabular}[c]{@{}c@{}}Size of\\Input\end{tabular} & 10 & 10 & 10 & 10 & 10 \\ \hline
\#Time Steps & 25 & 25 & 25 & 25 & 25 \\ \hline
\begin{tabular}[c]{@{}c@{}}Additional\\Details\end{tabular} &  &  & \begin{tabular}[c]{@{}c@{}}Pruning starts \\ at Epoch \#15k \\ and ends \\ at Epoch \#80k\end{tabular} &  &  \\ \hline
\begin{tabular}[c]{@{}c@{}}Size (KB)\\ assuming \\32 bit weights\end{tabular} & 243.42 & 30.92 & 31.02 & 30.86 & 26.38  \\ \hline
\begin{tabular}[c]{@{}c@{}}Mean\\Accuracy\end{tabular} & - & 88.69 & 87.25 & 91.26 & 91.66 \\ \hline
\begin{tabular}[c]{@{}c@{}}Top\\Accuracy\end{tabular} & 92.50 & 89.80 & 87.49 & 91.40 & 91.75 \\ \hline
\begin{tabular}[c]{@{}c@{}}Std\\Dev\\(Accuracy)\end{tabular} & - & 0.67 & 0.16 & 0.12 & 0.07 \\ \hline
\begin{tabular}[c]{@{}c@{}}Runtime\\(ms)\end{tabular} & 26.8 & 3.2 & 11.26 & 6.99 & 14 \\ \hline
\end{tabular}
\end{table*}

\begin{table*}
\centering
\caption{Hyperparameters for KWS-LSTM baseline network, network with LSTM layers compressed using HKPLSTM by a factor of 20 and equivalent sized networks compressed using LMF, Pruning and Small Baseline. LRD1 refers to Algorithm \ref{alg:lrd1}.}
\label{tab:hp-kwslstmhkp20}
\begin{tabular}{|c|c|c|c|c|c|}
\hline
Network & Baseline & Small Baseline & Pruning & LMF & HKPLSTM \\ \hline
\begin{tabular}[c]{@{}c@{}}Batch\\Size\end{tabular} & 100 & 100 & 100 & 100 & 100 \\ \hline
Optimizer & Adam & Adam & Adam & Adam & Adam \\ \hline
\begin{tabular}[c]{@{}c@{}}Weight\\Init\end{tabular} & \multicolumn{5}{c|}{glorot\_uniform} \\ \hline
\#Epochs & 30k & 80k & 100k & 80k & 90k \\ \hline
\begin{tabular}[c]{@{}c@{}}Initial\\   \\     LR\end{tabular} & 5x10\textasciicircum{}-4 & 10\textasciicircum{}-2 & 10\textasciicircum{}-2 & 10\textasciicircum{}-2 & 10\textasciicircum{}-2 \\ \hline
\begin{tabular}[c]{@{}c@{}}Decay\\Schedule\end{tabular} & \begin{tabular}[c]{@{}c@{}}5x10\textasciicircum{}-4,\\ 1x10\textasciicircum{}-4,\\ 2x10\textasciicircum{}-5 for\\ 10k steps each\end{tabular} & \begin{tabular}[c]{@{}c@{}}LRD1 \\     with \\ decay\_rate\\     0.09\end{tabular} & \begin{tabular}[c]{@{}c@{}}LRD1 \\     with \\ decay\_rate\\     0.09\end{tabular} & \begin{tabular}[c]{@{}c@{}}LRD1 \\     with \\ decay\_rate\\     0.09\end{tabular} & \begin{tabular}[c]{@{}c@{}}LRD1 \\     with \\ decay\_rate\\     0.09\end{tabular} \\ \hline
\#Layers & 1 & 1 & 1 & 1 & 1 \\ \hline
\begin{tabular}[c]{@{}c@{}}Hidden\\Vector\\Size\end{tabular} & 118 & 118 & 118 & 118 & 118 \\ \hline
\begin{tabular}[c]{@{}c@{}}Size of\\Input\end{tabular} & 10 & 10 & 10 & 10 & 10 \\ \hline
\#Time Steps & 25 & 25 & 25 & 25 & 25 \\ \hline
\begin{tabular}[c]{@{}c@{}}Additional\\Details\end{tabular} &  &  & \begin{tabular}[c]{@{}c@{}}Pruning starts \\ at Epoch \#15k \\ and ends \\ at Epoch \#80k\end{tabular} &  &  \\ \hline
\begin{tabular}[c]{@{}c@{}}Size (KB)\\ assuming \\32 bit weights\end{tabular} & 243.42 & 17.34 & 17.9 & 16.8 & 16.76  \\ \hline
\begin{tabular}[c]{@{}c@{}}Mean\\Accuracy\end{tabular} & - &  & 84.98 & 90.78 & 91.14 \\ \hline
\begin{tabular}[c]{@{}c@{}}Top\\Accuracy\end{tabular} & 92.50 & 89.80 & 85.17 & 90.9 & 91.28 \\ \hline
\begin{tabular}[c]{@{}c@{}}Std\\Dev\\(Accuracy)\end{tabular} & - & 0.58 & 0.16 & 0.11 & 0.11 \\ \hline
\begin{tabular}[c]{@{}c@{}}Runtime\\(ms)\end{tabular} & 26.8 & 2.25 & 8 & 5.78 & 15 \\ \hline
\end{tabular}
\end{table*}